\documentclass[english]{article}
\usepackage[T1]{fontenc}
\usepackage[latin9]{inputenc}
\usepackage{babel}
\usepackage{array}
\usepackage{multirow}
\usepackage{amsmath}
\usepackage{amsthm}
\usepackage{amssymb}
\usepackage{graphicx}
\usepackage{microtype}
\usepackage[unicode=true,pdfusetitle,
 bookmarks=true,bookmarksnumbered=false,bookmarksopen=false,
 breaklinks=false,pdfborder={0 0 0},pdfborderstyle={},backref=false,colorlinks=false]
 {hyperref}

\makeatletter

\usepackage[final]{neurips_2022}

\usepackage[table, svgnames, dvipsnames]{xcolor}
\usepackage{makecell, cellspace, caption, graphicx, colortbl}

\author{Kha Pham$^1$ \quad Hung Le$^1$ \quad Man Ngo$^2$ \quad Truyen Tran$^1$
	\\
	$^1$ Applied Artificial Intelligence Institute, Deakin University\\
	$^2$ Faculty of Mathematics and Computer Science, VNUHCM-University of Science
	\\\\
	$^1$ \texttt{\{phti, thai.le, truyen.tran\}@deakin.edu.au}\\
	$^2$ \texttt{nmman@hcmus.edu.vn}\\
}

\makeatother

\makeatother

\begin{document}
	\global\long\def\ModelName{\mathtt{FINE}}%
	
	\title{Functional Indirection Neural Estimator \\
		for Better Out-of-distribution Generalization}
	\maketitle
	\begin{abstract}
		The capacity to achieve out-of-distribution (OOD) generalization is
a hallmark of human intelligence and yet remains out of reach for
machines. This remarkable capability has been attributed to our abilities
to make conceptual abstraction and analogy, and to a mechanism known
as indirection, which binds two representations and uses one representation
to refer to the other. Inspired by these mechanisms, we hypothesize
that OOD generalization may be achieved by performing analogy-making
and indirection in the \emph{functional} space instead of the data
space as in current methods. To realize this, we design $\ModelName$
(Functional Indirection Neural Estimator), a neural framework that
\emph{learns to compose functions} that map data input to output on-the-fly.
$\ModelName$ consists of a backbone network and a trainable semantic
memory of basis weight matrices. Upon seeing a new input-output data
pair, $\ModelName$ dynamically constructs the backbone weights by
mixing the basis weights. The mixing coefficients are indirectly computed
through querying a separate corresponding semantic memory using the
data pair. We demonstrate empirically that $\ModelName$ can strongly
improve out-of-distribution generalization on IQ tasks that involve
geometric transformations. In particular, we train $\ModelName$ and
competing models on IQ tasks using images from the MNIST, Omniglot
and CIFAR100 datasets and test on tasks with unseen image classes
from one or different datasets and unseen transformation rules. $\ModelName$
not only achieves the best performance on all tasks but also is able
to adapt to small-scale data scenarios.

	\end{abstract}
	\vspace{-3mm}

	\section{\label{sec:Introduction} Introduction}
	
	\vspace{-3mm}
	
	\begin{center}
\emph{Every computer science problem can be solved with a higher level
of indirection.}
\par\end{center}

\begin{flushright}
---Andrew Koenig, Butler Lampson, David J. Wheeler
\par\end{flushright}

Generalizing to new circumstances is a hallmark of intelligence \cite{kriete2013indirection,biederman1988surface,geirhos2018generalisation}.
In some Intelligence Quotient (IQ) tests--a popular benchmark for
human intelligence--one must leverage their prior experience to identify
the hidden abstract rules out of a concrete example (e.g., a transformation
of an image) and then apply the rules to the next (e.g., a new set
of images of totally different appearance). These tasks necessitate
several key capabilities, including conceptual \emph{abstraction}
and \emph{analogy-making} \cite{mitchell2021abstraction}. Abstraction
allows us to extend a concept to novel situations. It is also driven
by analogy-making, which maps the current situation to the previous
experience stored in the memory. Indeed, analogy-making has been argued
to be one of the most important abilities of human cognition, or even
further, ``a concept is a package of analogies'' \cite{gentner2001analogical}.
The ability for humans to traverse seamlessly across concrete and
abstraction levels suggests another mechanism known as \emph{indirection}
to bind two representations and use one representation to refer to
the other \cite{kriete2013indirection,marcus2001algebraic}. 

\begin{figure}[th]
\begin{centering}
\includegraphics[scale=0.55]{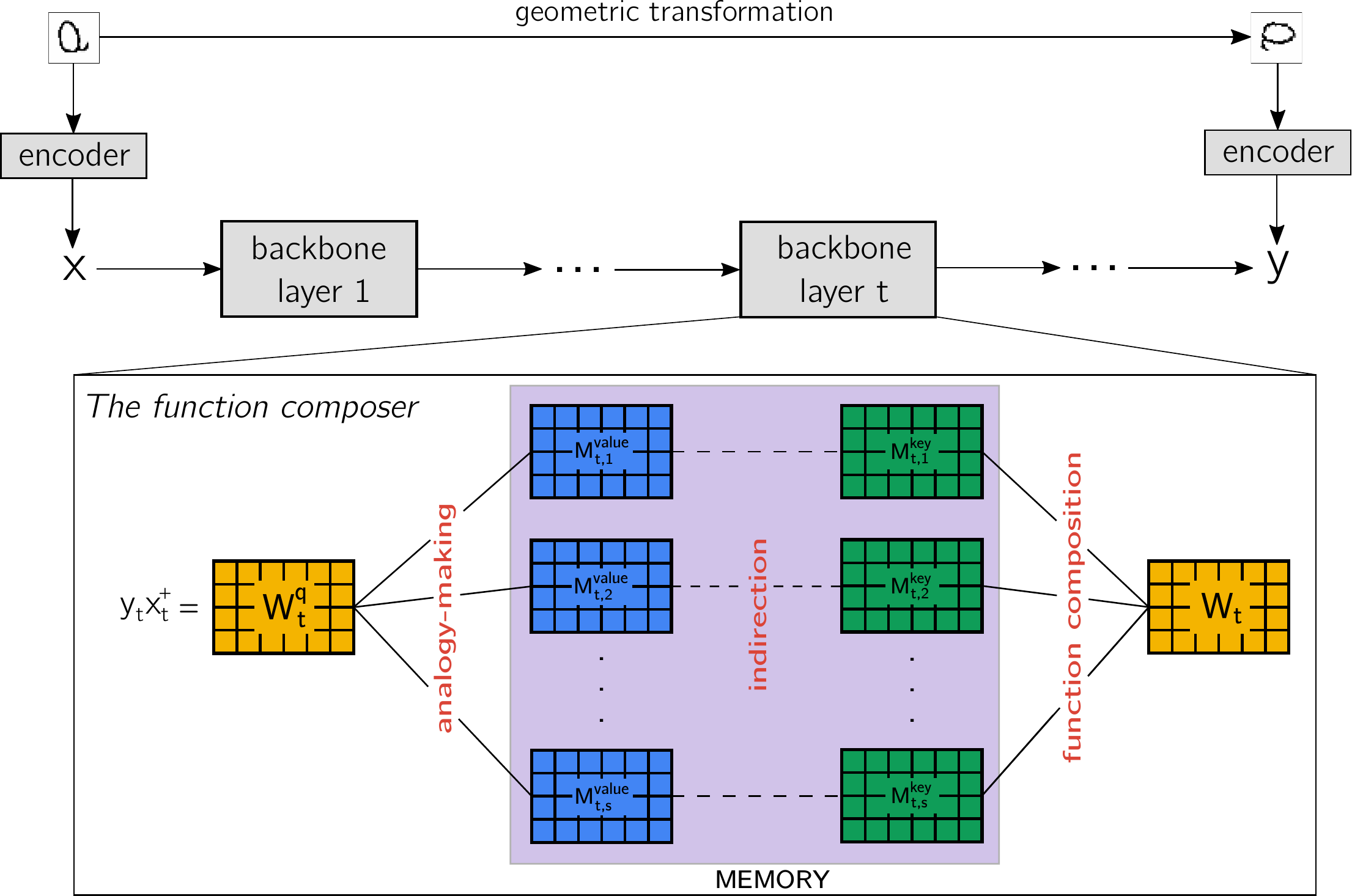}
\par\end{centering}
\caption{\label{fig:model-architecture} $\protect\ModelName$\emph{ }architecture.
\emph{Above: $\protect\ModelName$ }uses a pre-defined deep backbone
architecture to approximate a function mapping a given input embedding
$x$ to a given output embedding $y$\emph{. Below: }given the input
$x_{t}$ and pseudo-output $y_{t}$ of the $t^{\text{th}}$ backbone
layer, $\protect\ModelName$ first computes the query $W_{t}^{q}$
which represents the relation between $x_{t}$ and $y_{t}$. Then
$\protect\ModelName$ performs analogy-making to compare the query
with past experiences in the form of value memories. Finally, $\protect\ModelName$
binds the value memories with associated key memories via indirection
and computes the weight $W_{t}$ for the $t^{\text{th}}$ backbone
layer.}
\vspace{-3mm}
\end{figure}

Several deep learning models have successfully utilized analogy and
indirection. The Transformer \cite{Vaswani2017AttentionIA} and RelationNet
\cite{santoro2017simple} learn analogies between data, through self-attention
or pairwise functions. The ESBN \cite{Webb2021EmergentST} goes further
by incorporating the indirection mechanism to bind an entity to a
symbol and then reason on the symbols; and this has proved to be efficient
on tasks involving abstract rules, similar to those aforementioned
IQ tests.  However, a common drawback of these approaches is that
they operate on the data space, and thus are susceptible to out-of-distribution
samples.

In this paper, we propose to perform analogy-making and indirection
in \emph{functional} spaces instead. We aim to \emph{learn to compose}
a functional mapping from a given input to an output \emph{on-the-fly}.
By doing so, we achieve two clear advantages. First, since the class
of possible mappings is often restricted, it may not require a large
amount of training data to learn the distribution of functions. Second,
more importantly, since this approach performs indirection in functional
spaces, it avoids bindings between numerous entities and symbols in
data spaces, thus may help improve the out-of-distribution generalization
capability. 

To this end, we introduce a new class of problems that requires functional
analogy-making and indirection, which are deemed to be challenging
for current data-oriented approaches. The tasks are similar to popular
IQ tasks in which the model is given hints about the hidden rules,
then it has to predict the missing answer following the rules. One
reasonable approach is that models should be able to compare the current
task to what they saw previously to identify the rules between appearing
entities, and thus has to search on functional spaces instead of data
spaces. More concretely, we construct the IQ tasks by applying geometric
transformations to images from MNIST dataset, hand-written Omniglot
dataset and real-image CIFAR100 dataset, where the training set and
test set contain disjoint image classes from the same or different
datasets, and possibly disjoint transformation rules . 

Second, we present a novel framework named Functional Indirection
Neural Estimator (FINE) to solve this class of problems (see Fig.~\ref{fig:model-architecture}
for the overall architecture of $\ModelName$). $\ModelName$ consists
of (a) a neural backbone to approximate the functions and (b) a trainable
key-value memory module that stores the basis of the network weights
that spans the space of possible functions defined by the backbone.
The weight basis memories allow $\ModelName$ to perform analogy-making
and indirection in the function space. More concretely, when a new
IQ task arrives, $\ModelName$ first (1) takes the hint images to
make analogies with value memories, then (2) performs indirection
to bind value memories with key memories and finally (3) computes
the approximated functions based on key memories. Throughout a comprehensive
suite of experiments, we demonstrate that $\ModelName$ outperforms
the competing methods and adapts more effectively in small data scenarios.

	\section{\label{sec:Tasks} Tasks}
	
	For concreteness, we will focus on Intelligence Quotient (IQ) tests,
which have been widely accepted as one of the reliable benchmarks
to measure human intelligence \cite{rowe2012cognitive}. We will study
a popular class of IQ tasks that provides hints following some hidden
rules and requires the player to choose among given choices to fill
in a placeholder so that the filled-in entity obeys the rules of the
current task. In order to succeed in these tasks, the player must
be able to figure out the hidden rules and perform analogy-making
to select the correct choice. Moreover, once figuring out the rules
for the current task, a human player can almost always solve tasks
with similar rules regardless of the appearing entities given in the
tasks. This remarkable ability of out-of-distribution generalization
indicates that humans treat objects and relations abstractly instead
of relying on the raw sensory data. 

\begin{figure}
\begin{centering}
\includegraphics[scale=0.8]{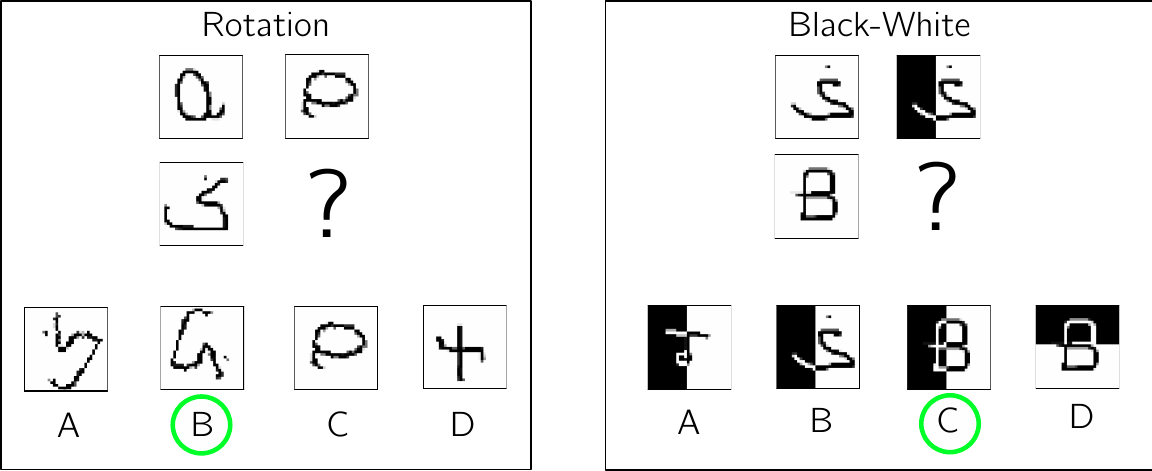}
\par\end{centering}
\caption{\label{fig:Tasks} Examples of two IQ tasks involving geometric transformations.
Choices with green circles are the correct solutions. \emph{Left.
}90-degree rotation. \emph{Right. }syntactic black-white transformation:
a part of the image is transformed to the opposite colors. }
\end{figure}

We aim to solve IQ tasks that involve geometric transformations (e.g.,
see Fig.~\ref{fig:Tasks} for an example), which include affine transformations
(translation, rotation, shear, scale, and reflection), non-linear
transformations (fisheye, horizontal wave) and syntactic transformations
(black-white, swap). Details of transformations are given in Supplementary.
In a task, the models are given 3 images $x,y$ and $x^{\prime}$,
where $y$ is the result of $x$ after applying a geometric transformation.
The models are then asked to select $y^{\prime}$ among 4 choices
$y_{1},y_{2},y_{3},y_{4}$ so that $(x^{\prime},y^{\prime})$ follows
the same rule as $(x,y)$ (i.e. if $y=f(x)$ then $y^{\prime}=f(x^{\prime})$
for transformation $f$). The 4 choices include (i) one with correct
object/image and correct transformation (which is the solution), (ii)
one with correct object/image and incorrect transformation, (iii)
one with incorrect object/image and correct transformation, and (iv)
one with both incorrect object/image and transformation. 

 Inspired by human capability, a reasonable approach to solve these
tasks is that models should be able to figure out the transformation
(or relation) between objects/images and apply the transformation
to novel objects/images. The datasets can be classified into two main
categories: single-transformation datasets and multi-transformation
datasets. Single-transformation datasets are ones that only include
a particular transformation, e.g. rotation. Note that the individual
transformations of the same type vary, e.g., rotations by different
angles. Multi-transformation datasets, on the other hand, consist
of several transformation types. To test the generalization capability
of the models, we build testing sets including classes of images that
have never been seen during training (see Section~\ref{subsec:Results-on-Omniglot}
and Section~\ref{subsec:Results-on-CIFAR100}), or even more challenging
tasks including unseen rules and unseen datasets (see Section~\ref{subsec:More-extreme-OOD}).
Models must be able to leverage knowledge and memory gained from the
training dataset to solve a new task. 

	\section{\label{sec:Method} Method}
	
	\subsection{\label{subsec:Functional-hypothesis} Functional Hypothesis}

Let $\mathcal{X}$ and $\mathcal{Y}$ be the data input and output
spaces, respectively. Denote by $\left(\mathcal{X}_{\text{train}},\mathcal{Y}_{\text{train}}\right)\subset\left(\mathcal{X},\mathcal{Y}\right)$
the training set, and $\left(\mathcal{X}_{\text{test}},\mathcal{Y}_{\text{test}}\right)\subset\left(\mathcal{X},\mathcal{Y}\right)$
the non-overlapping test set. Classical ML assumes that $\mathcal{X}_{\text{train}}$
and $\mathcal{X}_{\text{test}}$ are drawn from the same distribution.
Under this hypothesis, it is reasonable (for frequentists) to find
a function $f\colon\mathcal{X}\to\mathcal{Y}$ in the functional space
$\mathcal{F}$ that fits both the train and the test sets (i.e., the
discrepancy between $f(x)$ and corresponding $y$ is small for all
$(x,y)$ in $\left(\mathcal{X}_{\text{train}},\mathcal{Y}_{\text{train}}\right)$
and $\left(\mathcal{X}_{\text{test}},\mathcal{Y}_{\text{test}}\right)$).
However, when $\mathcal{X}_{\text{test}}$ is drawn from a different
data distribution from that of $\mathcal{X}_{\text{train}}$, it has
been widely reported that current deep learning models fail drastically
\cite{bahdanau2018systematic,csordas2021devil,greff2020binding,webb2020learning,Webb2021EmergentST}.
This is because the models are inferred exclusively from the observed
data distribution. Moreover, it could be the case that relations between
$x$ and $y$ in testing samples are unseen during training, which
raises questions on the feasibility of learning a single function
when dealing with out-of-distribution tasks. A natural solution for
this problem would be to train the model to learn the functions adaptively.
Formally speaking, the model will learn a \emph{function composer
}$\phi\colon\mathcal{X}\times\mathcal{Y}\to\mathcal{F}$ that maps
each pair of $(x,y)\in\mathcal{X}\times\mathcal{Y}$ (where $y$ is
the associated output of $x$) to a function $\phi_{x,y}$ in $\mathcal{F}$
so that $\phi_{x,y}$ approximates the true relation between $x$
and $y$. As discussed, training models this way leads to two clear
advantages: (1) it can help to handle the cases when there are multiple
(and possibly disjoint) relations between inputs and outputs within
the training and testing datasets; and (2) models are less dependent
on data and thus can achieve more stable results on different training
and testing sets. Empirical evidence for these points will be given
in Section~\ref{sec:Experiments}.

Since there are an infinite number of functions that can map an input
to an output with any given degree of precision, the key is to design
$\phi$ so that it can map a given input-output pair to a ``good
enough'' function. We humans may draw analogies between the current
situation and our experiences and then work out the most suitable
options \cite{gentner2001analogical}. For example, we  can recognize
a math problem in exam to be similar to a previous exercise in class
with different variable names. We may find the presented idea in a
paper related to that in another paper we read before, as someone
even says ``new ideas are just re-distribution of old ideas''. 
All these examples illustrate that analogy-making is a powerful strategy
in human thinking process. Inspired by this mechanism, we equip the
function composer $\phi$ with a semantic memory to store past knowledge,
on which analogy-making is performed. The memory also plays the role
of constraining the searching region for $\phi$, so that $\phi$
only looks for functions in the subspaces spanned by the memories
instead of the whole functional space $\mathcal{F}$. Further analysis
is given in Supplementary. 

A remaining question is how to design the memory. Let us be inspired
again by human thinking process: When we see an animal, we compare
its face, legs or tail with things we know and finally conclude it's
a ``dog''. Here ``dog'' is an abstract concept bound with primary
characteristics (e.g. face, legs, tail, etc.); once a new entity arrives,
we compare its properties with these characteristics, and if they
are similar, we utilize the indirection mechanism to infer it is indeed
the concept we are considering. In our case, the concepts are functions,
or more specifically, the geometric transformations. We thus maintain
a key-value memory structure, in which the keys represent abstract
concepts and values the associated characteristics of the concepts.
A new input-output pair matches with some values, and the indirection
enables us to compute the functions based on corresponding keys.

Note that we have swapped the role of keys and values of the traditional
memory (where the query is matched against the key, not value). This
is to emphasize that we perform indirection to map concrete functional
values to abstract functional keys.

	\subsection{\label{subsec:Model} Functional Indirection Neural Estimator}
	
	In this section, we present the Functional Indirection Neural Estimator
(FINE), an neural architecture to realize the general idea laid out
in Section~\ref{subsec:Functional-hypothesis}. $\ModelName$ learns
to compose a function mapping an input embedding $x$ to an output
embedding $y$ on-the-fly. The function is drawn from a parametric
family specified by a backbone neural network. Coupled with the backbone
is a \emph{function composer} $\phi$, which is trained to compute
the parameters of the backbone. More specifically, $\phi$ maps a
data input-output pair to the weights of the neural networks.

$\ModelName$ solves the proposed IQ tasks as follows: (1) Given two
images $x,y$ from the hint, $\ModelName$ uses $\phi$ to produce
the function transforming $x$ to $y$; denoted by $\phi_{x,y}$.
(2) Then, we feed the third image $x^{\prime}$ to $\phi_{x,y}$ and
get output $y^{*}=\phi_{x,y}(x')$. (3) We define a \emph{similarity
metric} to compare $y^{*}$ with given choices. The choice with the
closest distance from $y^{\ast}$ is model's answer. The function
composer and the similarity metric are specified as follows. 

\vspace{1mm}

\noindent\textbf{Encoder}: Images are encoded by a trainable encoder
(see Fig.~\ref{fig:model-architecture}). To effectively solve IQ
tasks introduced in Section~\ref{sec:Tasks}, we use the $p4$-CNN
\cite{cohen2016group} encoder to serve as an inductive bias for geometric
transformations. Throughout this paper, we refer to the images by
their embeddings. 

\subsubsection*{The functional memory}

Let us focus a weight matrix $W_{t}\in\mathbb{R}^{d_{t}^{\text{in}}\times d_{t}^{\text{out}}}$
at the $t$-th layer of the backbone network. In practice, $W_{t}$
belongs to the huge $d_{t}^{\text{in}}\times d_{t}^{\text{out}}$-dimensional
real matrix space $\mathcal{M}_{d_{t}^{\text{in}},d_{t}^{\text{out}}}$.
To reduce the complexity of $W_{t}$, we assume that $W_{t}$ only
belongs to a $s$-dimensional subspace of $\mathcal{M}_{d_{t}^{\text{in}},d_{t}^{\text{out}}}$,
where $s\ll d_{t}^{\text{in}}d_{t}^{\text{out}}$. This subspace has
a basis of $s$ matrices which are trainable and stored in $\ModelName$'s
memory, and $W_{t}$ will be written as a linear combination of these
matrices.

Denote by $M_{t}$ the memory for the $t$-th backbone layer. $M_{t}$
includes two sub-memories: the key memory $M_{t}^{\text{key}}=\{M_{t,1}^{\text{key}},\ldots,M_{t,s}^{\text{key}}\}$
and the corresponding value memory $M_{t}^{\text{value}}=\{M_{t,1}^{\text{value}},\ldots,M_{t,s}^{\text{value}}\}$.
Elements of $M_{t}^{\text{key}}$ and $M_{t}^{\text{value}}$ are
trainable matrices of size $d_{t}^{\text{in}}\times d_{t}^{\text{out}}$.
We further let $x_{t}\in\mathbb{R}^{d_{t}^{\text{in}}}$ and $y_{t}\in\mathbb{R}^{d_{t}^{\text{out}}}$
be the associated input and output, respectively, where $x_{t}$ is
output of the $(t-1)$-th layer and $y_{t}$ is the pseudo-output
computed by a trainable 1-layer neural network $y_{t}=\gamma_{t}(y)$. 

With this design, we control the complexity of functional hypothesis
space by either constraining the form of the backbone or the capacity
of the functional memory. 

\subsubsection*{Memory reading}

By the virtue of simplicity, we aim to find a simple query that can
demonstrate the relation between $x_{t}$ and $y_{t}$. Although the
exact relation may be non-linear, we found that a query induced from
linear relation is enough to efficiently read from memory. Formally,
we want to find a query $W_{t}^{q}$ such that $W_{t}^{q}x_{t}=y_{t}$.
The best-approximated solution is $W_{t}^{q}=y_{t}x_{t}^{+}$, where
$x_{t}^{+}$ is the pseudo-inverse of $x_{t}$ and can be efficiently
approximated by the iterative Ben-Israel and Cohen algorithm \cite{BenIsrael1966OnIC}.
This way of query computing requires no parameter as opposed to other
methods, where the input is often fed into a trainable neural network
to compute the query. 

With the query in hand, the next step is to perform analogy-making.
In $\ModelName$, the query $W_{t}^{q}$ represents for the current
situation and the value memory $M_{t}^{\text{value}}$ consists of
past experiences. The concrete query interacts with value memories
to measure how close the current situation is to each of the experiences.
The similarities are computed as dot products between the query and
value memories and normalized by a factor of $\sqrt{d_{t}^{\text{in}}d_{t}^{\text{out}}}$: 

\vspace{-3mm}
\begin{equation}
a_{t}=\frac{\text{fconcat}(M_{t}^{\text{value}})^{\top}\cdot\text{flatten}(W_{t}^{q})}{\sqrt{d_{t}^{\text{in}}d_{t}^{\text{out}}}},\label{eq:analogy-making}
\end{equation}

\vspace{-3mm}
where the \texttt{flatten} operator flattens the matrix $W_{t}^{q}$
of size $d_{t}^{\text{in}}\times d_{t}^{\text{out}}$ into a vector
of size $d_{t}^{\text{in}}d_{t}^{\text{out}}$, while the \texttt{fconcat}
operator first flattens all matrices in $M_{t}^{\text{value}}$, then
concatenates them together to form the value matrix of size $(d_{t}^{\text{in}}d_{t}^{\text{out}})\times s$.
The resulting \texttt{$a_{t}$} is a $s$-dimensional vector measuring
the similarities between the query and the entries in the value memory.
Here we omit the softmax operator as in usual attention to allow the
similarities with more freedom. The same idea is shared in the ESBN
\cite{Webb2021EmergentST}, where the softmax similarities are scaled
by a sigmoidal factor. 

Finally, the value memories are bound with their associated key memories
via indirection. This can be understood as moving forward from the
concrete space of data and value memories to the abstract space of
functions and key memories. With the key memories and the similarity
vector $a_{t}$, the weight $W_{t}$ of current backbone layer can
be computed as the linear combination of key memories: 

\vspace{-3mm}
\begin{equation}
W_{t}=\text{reshape}\left(\text{fconcat}(M_{t}^{\text{key}})\cdot a_{t}\right),\label{eq:function-composition}
\end{equation}

\vspace{-3mm}
where the \texttt{reshape} operator reshapes the vector of size $d_{t}^{\text{in}}d_{t}^{\text{out}}$
to a matrix of size $d_{t}^{\text{in}}\times d_{t}^{\text{out}}$.
Since the softmax is omitted when calculating the similarities, $W_{t}$
is not constrained to be in the convex hull of the key memories and
indeed can lie anywhere in the subspace spanned by those key memories. 

\vspace{-3mm}

\subsubsection*{Memory update}

The key and value memories are updated using gradient descent: 

\vspace{-3mm}
\begin{equation}
M_{t,i}^{\text{key/value}}\leftarrow M_{t,i}^{\text{key/value}}-\lambda\frac{\partial L}{\partial M_{t,i}^{\text{key/value}}},\quad\forall i=1,2,\ldots,s,\label{eq:memory-update}
\end{equation}

\vspace{-3mm}
where $\lambda>0$ is the learning rate and $L$ is the loss of the
training step.

\subsubsection*{The similarity metric}

After determining $\phi_{x,y}$, the model is given a new input $x^{\prime}$
and being asked to select the correct associated output $y^{\prime}$
among 4 choices $y_{1}^{\prime},y_{2}^{\prime},y_{3}^{\prime},y_{4}^{\prime}$
so that $(x^{\prime},y^{\prime})$ follows the same transformation
rule as $(x,y)$. This problem can be cast as finding the choice that
is the most similar with $y^{*}=\phi_{x,y}(x^{\prime})$. We consider
the weighted Euclidean metric that measures the distance between two
vectors $u,v\in\mathbb{R}^{d}$:

\vspace{-3mm}
\[
\eta(u,v)=\sum_{i=1}^{d}\alpha_{i}(u_{i}-v_{i})^{2},
\]

\vspace{-3mm}
where $\left\{ \alpha_{i}\right\} _{i=1}^{d}\ge0$ are trainable scalars,
i.e., each component of $u$ and $v$ contributes with different importance.
Finally, the probability to pick a choice is computed as: 

\vspace{-3mm}
\[
p\left(y_{i}^{\prime}\mid x^{\prime}\right)=\frac{\exp(-\eta(y_{i}^{\prime},y^{*}))}{{\displaystyle \sum_{j=1}^{4}\exp(}-\eta(y_{j}^{\prime},y^{*}))},\quad\text{for}\quad i=1,2,3,4.
\]

	\section{\label{sec:Experiments} Experiments}
	
	We conduct experiments to show the out-of-distribution generalization
capability of $\ModelName$ when performing tasks introduced in Section~\ref{sec:Tasks}.
For non-invertible transformations (e.g. reflection), we use a simple
2-layer MLP as the backbone. For invertible transformations, we use
the NICE architecture \cite{dinh2014nice} as backbone to serve as
an inductive bias for invertibility. Since in each NICE layer, only
half of the input is transformed, we use the same memory for two consecutive
layers, i.e., $M_{2t}=M_{2t+1}$. To balance between the backbone
complexity and computational cost, we use 4 NICE layers in all experiments.

We compare $\ModelName$ with three major classes of models: (a) models
that make analogies in the data space, including Transformer \cite{Vaswani2017AttentionIA},
PrediNet \cite{shanahan2020explicitly} and RelationNet \cite{santoro2017simple};
(b) models that leverage indirection to bind feature vectors with
associated symbols and reason on the symbols, including the ESBN \cite{Webb2021EmergentST};
and (c) models that aim to learn a mapping from data to the functional
space, including the HyperNetworks \cite{ha2016hypernetworks}. For
HyperNetworks, we still use the NICE backbone and just apply their
fast-weight generation method for fair comparisons. Except for $\ModelName$
and HyperNetworks, all models are trained with context normalization
\cite{webb2020learning}, which has been proved to be effective in
improving the generalization ability.

\textbf{Datasets \& implementation}: We generate data for IQ tasks
described in Section~\ref{sec:Tasks} using images from Omniglot
dataset \cite{Burda2016ImportanceWA}, which includes 1,623 handwritten
characters, and real-image CIFAR100 dataset \cite{Krizhevsky2009LearningML}.
If not specified, models are trained with $p4$-CNN encoder \cite{cohen2016group}.
Experiments are conducted using PyTorch on a single GPU with Adam
optimizer. Reported results are averages of 10 runs. 

\subsection{\label{subsec:Results-on-Omniglot} Results on Omniglot Dataset}

\begin{table}[th]
\begin{centering}
{\scriptsize{}}%
\begin{tabular}{l|>{\centering}p{0.6cm}|>{\centering}p{0.6cm}|>{\centering}p{0.6cm}|>{\centering}p{0.6cm}|>{\centering}p{0.6cm}|>{\centering}p{0.6cm}|>{\centering}p{0.9cm}|>{\centering}p{0.6cm}|>{\centering}p{0.6cm}|>{\centering}p{1cm}}
\cline{2-11} \cline{3-11} \cline{4-11} \cline{5-11} \cline{6-11} \cline{7-11} \cline{8-11} \cline{9-11} \cline{10-11} \cline{11-11} 
 & \multicolumn{9}{c|}{{\footnotesize{}Single-transformation}} & \multirow{3}{1cm}{\centering{}{\footnotesize{}Multi affine}}\tabularnewline
\cline{2-10} \cline{3-10} \cline{4-10} \cline{5-10} \cline{6-10} \cline{7-10} \cline{8-10} \cline{9-10} \cline{10-10} 
 & \multicolumn{5}{c|}{{\footnotesize{}Affine}} & \multicolumn{2}{c|}{{\footnotesize{}Non-linear}} & \multicolumn{2}{c|}{{\footnotesize{}Syntactic}} & \tabularnewline
\cline{2-10} \cline{3-10} \cline{4-10} \cline{5-10} \cline{6-10} \cline{7-10} \cline{8-10} \cline{9-10} \cline{10-10} 
 & \centering{}{\footnotesize{}Trans.} & \centering{}{\footnotesize{}Rot.} & \centering{}{\footnotesize{}Refl.} & \centering{}{\footnotesize{}Shear} & \centering{}{\footnotesize{}Scale} & \centering{}{\footnotesize{}Fish.} & \centering{}{\footnotesize{}H.Wave} & \centering{}{\footnotesize{}B\&W} & \centering{}{\footnotesize{}Swap} & \tabularnewline
\hline 
{\footnotesize{}RelationNet} & {\footnotesize{}27.1} & {\footnotesize{}26.2} & {\footnotesize{}25.5} & {\footnotesize{}27} & {\footnotesize{}27.5} & {\footnotesize{}26.1} & {\footnotesize{}30.2} & {\footnotesize{}49.7} & {\footnotesize{}26.0} & {\footnotesize{}25.3}\tabularnewline
{\footnotesize{}PrediNet} & {\footnotesize{}68.5} & {\footnotesize{}43.9} & {\footnotesize{}32.9} & {\footnotesize{}62.4} & {\footnotesize{}65.7} & {\footnotesize{}36.2} & {\footnotesize{}46.1} & {\footnotesize{}60.5} & {\footnotesize{}57.5} & {\footnotesize{}34.9}\tabularnewline
{\footnotesize{}HyperNet} & {\footnotesize{}88.9} & {\footnotesize{}62.0} & {\footnotesize{}94.0} & {\footnotesize{}74.5} & {\footnotesize{}81.8} & {\footnotesize{}63.2} & {\footnotesize{}80.4} & {\footnotesize{}88.6} & {\footnotesize{}90.1} & {\footnotesize{}54.0}\tabularnewline
{\footnotesize{}Transformer} & {\footnotesize{}89.5} & {\footnotesize{}64.8} & {\footnotesize{}44.3} & {\footnotesize{}86.3} & {\footnotesize{}84.0} & {\footnotesize{}41.4} & {\footnotesize{}91.0} & {\footnotesize{}97.6} & {\footnotesize{}49.9} & {\footnotesize{}59.4}\tabularnewline
{\footnotesize{}ESBN} & {\footnotesize{}79.8} & {\footnotesize{}58.6} & {\footnotesize{}50.1} & {\footnotesize{}84.5} & {\footnotesize{}83.4} & {\footnotesize{}67.1} & {\footnotesize{}86.4} & {\footnotesize{}90.5} & {\footnotesize{}71.6} & {\footnotesize{}63.1}\tabularnewline
\hline 
{\footnotesize{}$\ModelName$-MLP} & \textbf{\footnotesize{}96.1} & {\footnotesize{}73.9} & \textbf{\footnotesize{}95.1} & {\footnotesize{}84.5} & {\footnotesize{}86.0} & {\footnotesize{}70.4} & {\footnotesize{}84.8} & {\footnotesize{}94.9} & {\footnotesize{}87.5} & \textbf{\footnotesize{}72.0}\tabularnewline
{\footnotesize{}$\ModelName$-NICE} & {\footnotesize{}94.3} & \textbf{\footnotesize{}77.6} & {\footnotesize{}57.7} & \textbf{\footnotesize{}87.2} & \textbf{\footnotesize{}86.6} & \textbf{\footnotesize{}78.5} & \textbf{\footnotesize{}95.9} & \textbf{\footnotesize{}98.4} & \textbf{\footnotesize{}96.2} & {\footnotesize{}69.1}\tabularnewline
\hline 
\end{tabular}{\scriptsize\par}
\par\end{centering}
\caption{\label{tab:Omniglot-transformation-task}  Test accuracy (\%) on
Omniglot dataset. }
\end{table}

We use 100 characters for training and other 800 characters for testing.
The train and test set size is 10,000 and 20,000, respectively. For
$\ModelName$, we use 4 NICE layers with 48 memories for each pair
of NICE layers. Experimental results for single-transformation tasks
are shown in Table~\ref{tab:Omniglot-transformation-task}. Overall,
$\ModelName$ dominates others with large margins. For example, the
gap to the runner-up on the Rotation task is nearly 13\%. With test
accuracy over $75\%$ on all tasks, $\ModelName$ shows a strong capability
of out-of-distribution generalization. 

We further conduct multi-affine-transformation experiments. In this
case, the training set includes multiple types of affine transformations,
while other settings are similar to the single-transformation case.
Results are also reported in Table~\ref{tab:Omniglot-transformation-task}.
$\ModelName$ continues to outperform other models. This is because
only $\ModelName$ explicitly assumes the existence of multiple good
functions that represent the transformation from input to output data.
We note that although HyperNet also makes a similar assumption by
generating data-specific weights, it does not utilize analogy-making
and indirection and thus, fails to generalize to unseen images. 

\subsection{\label{subsec:Results-on-CIFAR100} Results on CIFAR100 Dataset }

\begin{table}[th]
\begin{centering}
\begin{tabular}{l|>{\centering}m{0.9cm}|>{\centering}m{0.9cm}|>{\centering}m{0.9cm}|>{\centering}m{0.9cm}|>{\centering}m{0.9cm}|>{\centering}m{0.9cm}|>{\centering}m{0.9cm}|>{\centering}m{0.9cm}|>{\centering}m{0.9cm}}
\cline{2-10} \cline{3-10} \cline{4-10} \cline{5-10} \cline{6-10} \cline{7-10} \cline{8-10} \cline{9-10} \cline{10-10} 
\multicolumn{1}{l}{} & \multicolumn{5}{c|}{Affine} & \multicolumn{2}{c|}{Non-linear} & \multicolumn{2}{c}{Syntactic}\tabularnewline
\cline{2-10} \cline{3-10} \cline{4-10} \cline{5-10} \cline{6-10} \cline{7-10} \cline{8-10} \cline{9-10} \cline{10-10} 
\multicolumn{1}{l}{} & Trans. & Rot. & Refl. & Shear & Scale & Fish. & H.Wave  & B\&W & Swap\tabularnewline
\hline 
RelationNet & 59.9 & 49.6 & 29.9 & 45.3 & 66.2 & 28.7 & 39.5 & 26.2 & 29.7\tabularnewline
PrediNet & 72.4 & 65.6 & 40.6 & 74.3 & 76.1 & 37.1 & 53.9 & 32.7 & 39.6\tabularnewline
HyperNet & 94.8 & 86.8 & 46.6 & 91.3 & 85.2 & 46.8 & 80.5 & 47.8 & 46.0\tabularnewline
Transformer & 98.4 & 86.3 & 47.5 & 95.4 & 84.9 & 47.2 & 95.1 & 51.6 & 47.6\tabularnewline
ESBN & 96.6 & 81.9 & 50.6 & 90.1 & 81.5 & 57.7 & 95.7 & 68.8 & 50.5\tabularnewline
\hline 
$\ModelName${\small{}-MLP} & 98.9 & 89.7 & \textbf{80.6} & \textbf{95.7} & 86.8 & 59.6 & 95.2 & 83.1 & 50.8\tabularnewline
$\ModelName${\small{}-NICE} & \textbf{99.2} & \textbf{91.3} & 51.1 & 95.6 & \textbf{87.0} & \textbf{76.8} & \textbf{98.3} & \textbf{89.1} & \textbf{51.6}\tabularnewline
\hline 
\end{tabular}
\par\end{centering}
\caption{\label{tab:CIFAR100-single-transformation-task} Test accuracy (\%)
on CIFAR100 dataset of single-transformation tasks. For readability
we report only the mean values here. Full table is reported in Supplementary.}
\end{table}

\begin{table}[th]
\begin{centering}
\begin{tabular}{>{\raggedright}p{2.3cm}|>{\centering}p{1.5cm}>{\centering}p{1.5cm}>{\centering}p{1.5cm}>{\centering}p{1.5cm}|>{\centering}p{1.5cm}>{\centering}p{1.5cm}}
\hline 
 & RelationNet & PrediNet & Transformer & ESBN & $\ModelName$

{\small{}-MLP} & $\ModelName$

{\small{}-NICE}\tabularnewline
\hline 
Group CNN & $32.5\pm1.1$ & $46\pm1.0$ & $67.8\pm4.5$ & $71.1\pm0.5$ & $79.6\text{\ensuremath{\pm0.5}}$ & $\mathbf{81.6\pm0.5}$\tabularnewline
3-layer ResNet & $31.1\pm2.7$ & $55.9\pm1.7$ & $28.5\pm0.5$ & $66.8\pm9.7$ & $\mathbf{77.8\pm0.5}$ & $73.5\pm1.6$\tabularnewline
MLP & $47.2\pm2.6$ & $59.3\pm0.8$ & $61.5\pm1.4$ & $54.7\pm2.6$ & $\mathbf{72.9\pm1.1}$ & $70.0\pm1.3$\tabularnewline
\end{tabular}
\par\end{centering}
\caption{\label{tab:CIFAR100-multi-transformation-task} Test accuracy (\%)
on CIFAR100 dataset of multi-affine-transformation tasks.}
\end{table}

For CIFAR100 dataset, we follow similar settings as in experiments
on the Omniglot dataset, except that we use 50 classes for training
and 50 remaining classes for testing. We also conduct experiments
on single-transformation and multi-affine-transformation tasks. Results
for single-transformation tasks are reported in Table~\ref{tab:CIFAR100-single-transformation-task}.
Again, $\ModelName$ outperforms all other models on all tasks, especially
on Reflection where the gap is nearly 30\%. Although $\ModelName$
does not show good performance on Swap task as in Omniglot experiments,
it is still slightly better than other models. On the remaining tasks,
$\ModelName$ achieves test accuracy of more than 80\%.

For the multi-affine-transformation task, we report the performances
of the models when trained with different encoder architectures, including
the $p4$-CNN, 3-layer ResNet and 2-layer MLP, in Table~\ref{tab:CIFAR100-multi-transformation-task}.
The results show two superior characteristics of $\ModelName$: first,
$\ModelName$ is consistently better than other models across different
encoders; second, $\ModelName$ is more stable with small standard
deviations. This empirical result supports the functional hypothesis
stated in Section~\ref{subsec:Functional-hypothesis}, where we suggest
that focusing on functions distribution instead of data distribution
can boost model's generalization capability and stability.

\subsection{\label{subsec:More-extreme-OOD} More Extreme OOD Tasks}

\begin{table}[th]
\begin{centering}
{\footnotesize{}}%
\begin{tabular}{>{\raggedright}p{1.8cm}|>{\centering}m{0.9cm}|>{\centering}p{0.9cm}|>{\centering}p{0.9cm}|>{\centering}m{0.9cm}|>{\centering}p{0.9cm}|>{\centering}p{0.9cm}|>{\centering}m{0.9cm}|>{\centering}p{0.9cm}|>{\centering}p{0.9cm}}
\hline 
{\small{}Train set}{\small\par}

{\small{}Test set} & \multicolumn{3}{>{\raggedleft}p{2.4cm}|}{{\small{}CIFAR 100}{\small\par}

{\small{}CIFAR 100}} & \multicolumn{3}{>{\raggedleft}p{2.4cm}|}{{\small{}CIFAR100}{\small\par}

{\small{}Omniglot}} & \multicolumn{3}{>{\raggedleft}p{2.4cm}}{{\small{}Omniglot}{\small\par}

{\small{}MNIST}}\tabularnewline
\hline 
 & {\small{}Trans.} & {\small{}Rot.} & {\small{}Shear} & {\small{}Trans.} & {\small{}Rot.} & {\small{}Shear} & {\small{}Trans.} & {\small{}Rot.} & {\small{}Shear}\tabularnewline
\hline 
{\small{}RelationNet} & {\small{}25.4} & {\small{}26.5} & {\small{}26.4} & {\small{}25} & {\small{}24.9} & {\small{}25} & {\small{}24.8} & {\small{}25} & {\small{}25.2}\tabularnewline
{\small{}PrediNet} & {\small{}25.5} & {\small{}39.4} & {\small{}36.2} & {\small{}26.2} & {\small{}26.1} & {\small{}26.3} & {\small{}23.8} & {\small{}26.4} & {\small{}25.8}\tabularnewline
{\small{}HyperNet} & {\small{}31.6} & {\small{}67.2} & {\small{}58.6} & {\small{}22.2} & {\small{}26.4} & {\small{}22.9} & {\small{}22.7} & {\small{}29} & {\small{}28.2}\tabularnewline
{\small{}Transformer} & {\small{}30.9} & {\small{}64.2} & {\small{}55} & \textbf{\small{}30} & {\small{}31.7} & {\small{}33.8} & {\small{}28.4} & {\small{}27} & {\small{}25.1}\tabularnewline
{\small{}ESBN} & {\small{}16.4} & {\small{}81.3} & {\small{}42.8} & {\small{}15.4} & {\small{}39.2} & {\small{}33.3} & {\small{}17.7} & {\small{}41.3} & {\small{}36.6}\tabularnewline
\hline 
{\small{}$\ModelName$} & \textbf{\small{}62} & \textbf{\small{}85.6} & \textbf{\small{}77.8} & {\small{}22.1} & \textbf{\small{}43.2} & \textbf{\small{}39.2} & \textbf{\small{}39.4} & \textbf{\small{}44.7} & \textbf{\small{}37.9}\tabularnewline
\hline 
\end{tabular}{\footnotesize\par}
\par\end{centering}
\caption{\textbf{\label{tab:more-extreme-ood}}Test accuracy (\%) on more extreme
OOD tasks with unseen objects, unseen rules and (possibly) unseen
datasets. \emph{Translation: }train with translation vectors $(a,b)$
with $|a|,|b|\le3$, test with either $|a|>3$ or $|b|>3$. \emph{Rotation:
}train with rotation angles $\alpha\le180^{\circ}$, test with $\alpha>180^{\circ}$.
\emph{Shear: }train with shear angles $(\alpha,\beta)$ with $|\alpha|,|\beta|\le30^{\circ}$,
test with either $|\alpha|>30^{\circ}$ or $|\beta|>30^{\circ}$.}
\end{table}

Previous tasks only include unseen classes of objects during testing.
In this section, we further test $\ModelName$ and related models
on more challenging OOD tasks: tasks with unseen rules during training
and even ones with images from unseen datasets. The training and testing
sets are either images from CIFAR100, Omniglot or MNIST datasets,
while the hidden rules are either translation, rotation or shear.
For translation, training problems consist of translation vectors
$(a,b)$ with $|a|,|b|\le3$, and models are tested with either $|a|>3$
or $|b|>3$; for rotation, model are trained with angles $\alpha\le180^{\circ}$
and tested with $\alpha>180^{\circ}$; for shear, training angles
$(\alpha,\beta)$ are ones with $|\alpha|,|\beta|\le30^{\circ}$,
while testing ones are either $|\alpha|>30^{\circ}$ or $|\beta|>30^{\circ}$.
All tasks have 5,000 data points for training and 10,000 for testing.
We report results of $\ModelName$ with NICE backbone and related
models in Table \ref{tab:more-extreme-ood}. As expected, $\ModelName$
continues to outperform other models on most of the tasks, even on
extreme OOD tasks with unseen datasets and unseen rules where performances
of all models drop significantly. This demonstrates $\ModelName$
is capable of effectively learning the basis weights, which are stored
in the memory, to represent novel rules. 

\vspace{-3mm}

\subsection{Model Analysis and Ablation Study}

\subsubsection*{Clustering on functional spaces}

\begin{figure}[th]
\begin{centering}
\includegraphics[scale=0.7]{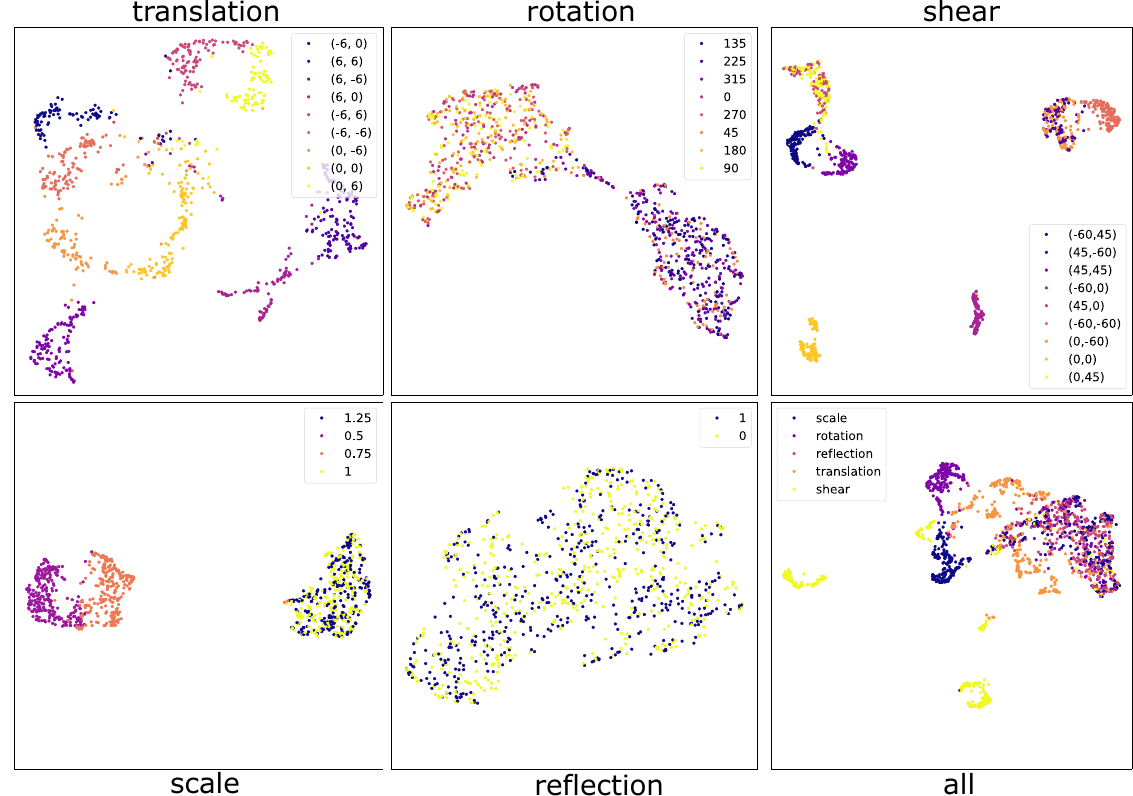}
\par\end{centering}
\caption{\label{fig:umap-clustering} \emph{Clusters on functional space}.
We project weights produced by the function composer $\phi$ to see
whether $\phi_{x_{1},y_{1}}$ locates closely to $\phi_{x_{2},y_{2}}$
on the functional space if the relation between $x_{1}$ and $y_{1}$
is similar to that between $x_{2}$ and $y_{2}$. Except for Reflection,
the weights are nicely clustered.}
\end{figure}

We study how the transformations are distributed in the functional
space. We use the $\ModelName$ model trained on multi-affine transformation
tasks. For an input-output pair $(x,y)$, we flatten and concatenate
all weights of NICE layers to form a vector representing $\phi_{x,y}$.
We then use the UMAP \cite{mcinnes2018umap} to project $\phi_{x,y}$'s
vectors onto the 2D plane. Results are shown in Fig.~\ref{fig:umap-clustering}.
It is interesting to see that shears with the same horizontal or vertical
angles are positioned closely; and scales are separated into 2 ``big''
clusters, one for smaller scale and one for bigger scale. In contrast,
reflection representations seem not to be clustered properly, which
is worth further investigation in future work.

\subsubsection*{Number of memories and backbone layers}

We do an ablation study to see the effect of number of memories and
backbone layers in $\ModelName$. For the limit case with 0 memory,
we assign the query matrices to be the weights for NICE layers without
the analogy-making and indirection process. For the limit case with
0 NICE layer, we replace the NICE backbone by a 2-layer MLP. 

\begin{figure}[th]
\begin{centering}
\includegraphics[scale=0.6]{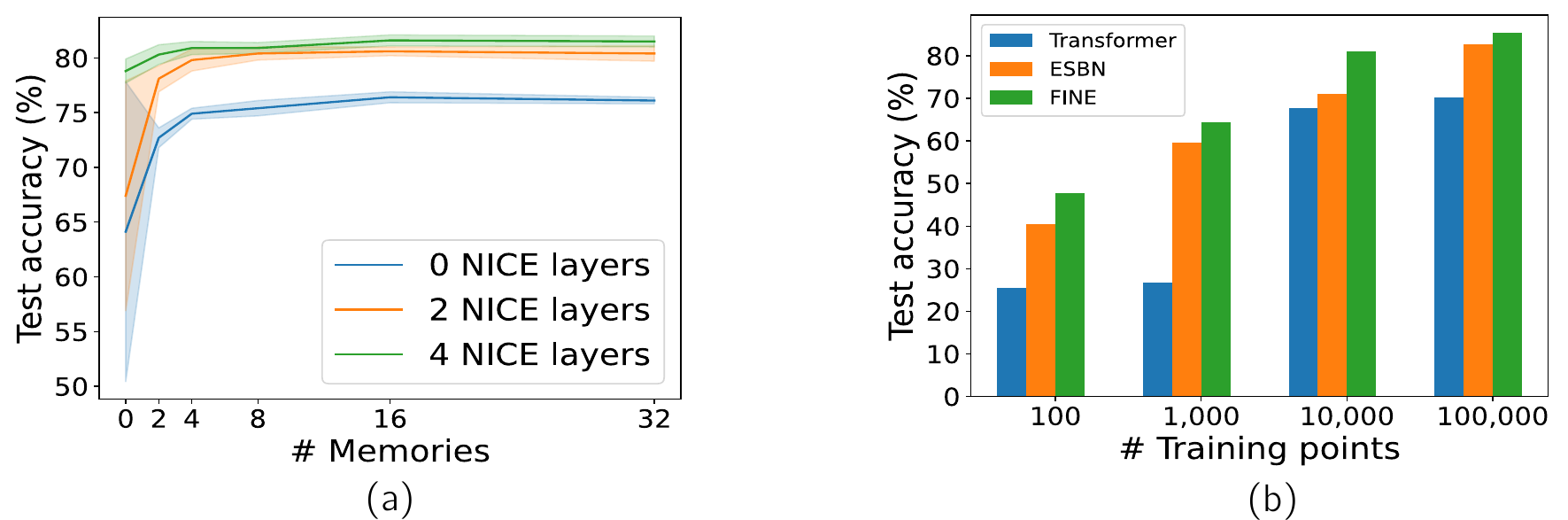}
\par\end{centering}
\caption{\label{fig:ablation-num-memory-training-points} (a) \emph{Performance
of $\protect\ModelName$ with different number of memories and backbone
layers}. Overall, the test accuracy increases with more memories and
backbone layers. (b) \emph{Performance with different number of training
points}. $\protect\ModelName$ works fairly well on small datasets
while others may fail significantly.}
\end{figure}

Results are shown in Fig.~\ref{fig:ablation-num-memory-training-points}(a).
Overall, we can observe clear improvements when we increase the number
of memories or number of NICE layers. More interestingly, the more
number of memories is, the more stable the results will be. Increasing
the number of memories is equivalent to enlarging the range of $\phi$,
and increasing the number of NICE layers is equivalent to enlarging
the hypothesis space $\mathcal{F}$.  Enlarging $\mathcal{F}$ may
help $\ModelName$ approach the true functions while still being sufficiently
constrained by the number of memories, thus still being able to maintain
its generalization capability. 

\subsubsection*{Number of training data points}

We train models on multi-affine-transformation task with training
sets of different sizes. Results are reported in Fig.~\ref{fig:ablation-num-memory-training-points}(b).
$\ModelName$ can adapt with small datasets of sizes 100 or 1000 and
achieve fair accuracy (47.8\% and 64.4\%, respectively). Moreover,
$\ModelName$ achieves average test accuracy of 81.1\% on training
set of size 10,000, which is quite close to the 85.3\% accuracy on
100,000 data points training sets. This shows $\ModelName$ may obtain
a near-optimal solution even with small number of training data points.
In contrast, ESBN and Transformer need 100,000 training data points
to achieve 80\% or higher test accuracy, while only achieving roughly
70\% test accuracy on smaller datasets. 

\subsubsection*{}

	\vspace{-6mm}

	\section{\label{sec:Related-work} Related work}
	
	In recent years, there has been a strong interest in designing models
that are capable of generalizing systematically. The Module Networks
\cite{andreas2016neural}, which dynamically compose neural networks
out of trainable modules, have been shown to possess some degree of
systematic generalization \cite{bahdanau2018systematic}. Parascandolo
et al. \cite{parascandolo2018learning} proposed a method to train
multiple competing experts to explain image transformations on MNIST
and Omniglot datasets, yet the transformations are simpler than ones
in our $\ModelName$ dataset and it is not clear whether the proposed
method can deal with unseen transformations. The Neural Interpreter
\cite{rahaman2021dynamic} uses attention mechanism to recompose functional
modules for each input-output pair and test their method on abstract
reasoning tasks, yet tends to require a large amount of data to learn.
Switch Transformer \cite{fedus2021switch} mitigates the communication
and computational cost in mixture of experts models.  Recently,
ESBN \cite{Webb2021EmergentST}, which utilizes the indirection mechanism,
shows a great promise on OOD tasks. Both methods achieve their degree
of systematic generalization by injecting symbolic biases into the
models. Our model $\ModelName$ follows the same strategy, but it
performs analogy-making and indirection in functional spaces instead
of data spaces, and this has proved to boost both the performance
and stability.

IQ tests are powerful testbeds for visual and abstract reasoning.
Inspired by Raven's Progressive Matrices (RPM), the RAVEN dataset
\cite{zhang2019raven} was proposed as a testbed for visual reasoning
models. However, this dataset does not focus on testing the ability
of OOD generalization. Webb et al. \cite{Webb2021EmergentST} propose
a series of IQ tasks with Unicode characters to show the effectiveness
of indirection in tasks involving abstract rules, however these tasks
are relatively simple since they only require the models to understand
the same-different relation. The ARC dataset \cite{chollet2019measure}
aims to serve as a benchmark for general intelligence and includes
various psychometric IQ tasks in the form of grid structures. In this
paper, we propose IQ tasks involving geometric transformations as
introduced in Section~\ref{sec:Tasks}. These tasks are not only
flexible so that we can include images from different datasets or
create/combine numerous transformations, but also challenging to test
OOD generalization abilities of models. 

The weight composition feature of $\ModelName$ links back to the
concept of fast weights \cite{malsburg1994correlation}, the idea
of computing data-specific network weights on-the-fly. HyperNetworks
\cite{ha2016hypernetworks} stylizes this idea by computing the fast
weights using a separate trainable (slow weight) network. The Meta-learned
Neural Memory \cite{Munkhdalai2019MetalearnedNM} uses the pseudo-target
technique (which we also leverage in $\ModelName$) and appropriately
updates the short-term memory once new input arrives. The Neural Stored-Program
Memory (NSM) \cite{le2020neural} proposes a hybrid approach between
slow-weight and fast-weight to compute network weights on-the-fly
based on slow-weight key and value memories. However, NSM only performs
on sequential learning tasks, while $\ModelName$ aims to solve OOD
IQ tasks requiring abstract cognition. Moreover, $\ModelName$ computes
the query based on the input and the (pseudo-) output, while NSM's
query is computed based on the input only. Memories have been to be
versatile in meta-learning and few-shot learning \cite{kaiser2017learning,santoro2016meta,vinyals2016matching}
due to the ability to rapidly store past examples and adapt to new
situations. In our case, an IQ task can be thought of as a one-shot
learning scenario in which $\ModelName$ has to make use of the long-term
key-value memory to adapt to the current task.

	\vspace{-3mm}

	\section{\label{sec:Limitations} Limitations}
	
	Operating in functional spaces requires $\ModelName$'s memory to
store several trainable weight matrices to infer the data-specific
weights for the backbone. Moreover, each layer of the backbone requires
its own memory, thus $\ModelName$ may need a large number of parameters
for very deep backbones. This could be addressed by parameter-sharing
across layers and limiting the rank of the weight matrices. 

It remains to design the backbone architecture optimally. We have
used the NICE architecture as backbone for invertible transformations
and a 2-layer MLP backbone for non-invertible cases. We further investigated
our model's performances with larger number of training points (up
to 1M) and observed that $\ModelName$ with the NICE backbone peaks
at 100,000 training points with around $85\%$ test accuracy, while
$\ModelName$ with MLP backbone continues to improve and achieves
around $95\%$ test accuracy at 1M training points. This calls for
further theoretical analysis to guide the architectural design of
the backbone network. 

Finally, testing $\ModelName$ on IQ tasks in the visual space may
limit its potential on IQ tasks involving other modalities. For example,
one may consider a textual IQ problem: ``if $\text{abc}\to\text{abd}$,
then $\text{mmnnpp}\to?$''. It is worth to emphasize that the concepts
of analogy-making and indirection in functional spaces are indeed
general and thus the idea of $\ModelName$ should be applicable to
various scenarios.

	\vspace{-3mm}

	\section{\label{sec:Discussion} Conclusion}
	
	To study the out-of-distribution (OOD) generalization capability of
models, we have proposed IQ tasks that require models to rapidly recognize
the hidden rules of geometric transformations between a pair of images
and transfer the rules to a new pair of different image classes. Such
tasks would necessitate human-like abilities for conceptual abstraction,
analogy-making, and utilizing indirection. We put forward a hypothesis
that these mechanisms should be performed in the functional space
instead of data space as in current deep learning models. To realize
the hypothesis, we then proposed $\ModelName$, a memory-augmented
neural architecture that learns to compose functions mapping an input
to an output on-the-fly. The memory has two trainable components:
the value sub-memory and the binding key sub-memory, where the keys
are basis weight matrices that span the space of functions. For an
IQ task, when given a hint in the form of an input-output pair, and
$\ModelName$ estimates the analogy between the pair and the values
as mixing coefficients. These coefficients are then used to mix the
binding keys via indirection to generate the weights of the backbone
neural net which computes the intended function. For a test input
of different class, the function is used to estimate the most compatible
output, thus solving the IO task. Through an extensive suite of experiments
using images from the Omniglot and CIFAR100 datasets to construct
the IQ tasks, $\ModelName$ is found to be reliable in figuring out
the hidden relational pattern in each IQ task and thus is able to
solve new tasks, even with unseen image classes. Importantly, $\ModelName$
outperforms other models in all experiments, and can generalize well
in small data regimes.

Future works will include making $\ModelName$ robust against OOD
in transformations without catastrophic forgetting when new transformations
are continually introduced.

	\newpage
	
	\bibliographystyle{plain}

\begin{thebibliography}{10}

\bibitem{andreas2016neural}
Jacob Andreas, Marcus Rohrbach, Trevor Darrell, and Dan Klein.
\newblock Neural module networks.
\newblock In {\em Proceedings of the IEEE conference on computer vision and
  pattern recognition}, pages 39--48, 2016.

\bibitem{bahdanau2018systematic}
Dzmitry Bahdanau, Shikhar Murty, Michael Noukhovitch, Thien~Huu Nguyen, Harm
  de~Vries, and Aaron Courville.
\newblock Systematic generalization: What is required and can it be learned?
\newblock In {\em International Conference on Learning Representations}, 2018.

\bibitem{BenIsrael1966OnIC}
A.~Ben-Israel and D.~Cohen.
\newblock On iterative computation of generalized inverses and associated
  projections.
\newblock {\em SIAM Journal on Numerical Analysis}, 3:410--419, 1966.

\bibitem{biederman1988surface}
Irving Biederman and Ginny Ju.
\newblock Surface versus edge-based determinants of visual recognition.
\newblock {\em Cognitive psychology}, 20(1):38--64, 1988.

\bibitem{Burda2016ImportanceWA}
Yuri Burda, Roger~B. Grosse, and R.~Salakhutdinov.
\newblock Importance weighted autoencoders.
\newblock {\em ICLR}, 2016.

\bibitem{chollet2019measure}
Fran{\c{c}}ois Chollet.
\newblock On the measure of intelligence.
\newblock {\em arXiv preprint arXiv:1911.01547}, 2019.

\bibitem{cohen2016group}
Taco Cohen and Max Welling.
\newblock Group equivariant convolutional networks.
\newblock In {\em International conference on machine learning}, pages
  2990--2999. PMLR, 2016.

\bibitem{csordas2021devil}
R{\'o}bert Csord{\'a}s, Kazuki Irie, and J{\"u}rgen Schmidhuber.
\newblock The devil is in the detail: Simple tricks improve systematic
  generalization of transformers.
\newblock {\em arXiv preprint arXiv:2108.12284}, 2021.

\bibitem{dinh2014nice}
Laurent Dinh, David Krueger, and Yoshua Bengio.
\newblock {NICE: Non-linear independent components estimation}.
\newblock {\em arXiv preprint arXiv:1410.8516}, 2014.

\bibitem{fedus2021switch}
William Fedus, Barret Zoph, and Noam Shazeer.
\newblock Switch transformers: Scaling to trillion parameter models with simple
  and efficient sparsity, 2021.

\bibitem{geirhos2018generalisation}
Robert Geirhos, Carlos~RM Temme, Jonas Rauber, Heiko~H Sch{\"u}tt, Matthias
  Bethge, and Felix~A Wichmann.
\newblock Generalisation in humans and deep neural networks.
\newblock {\em Advances in neural information processing systems}, 31, 2018.

\bibitem{gentner2001analogical}
Dedre Gentner, Keith~J Holyoak, and Boicho~N Kokinov.
\newblock {\em The analogical mind: Perspectives from cognitive science}.
\newblock MIT press, 2001.

\bibitem{greff2020binding}
Klaus Greff, Sjoerd Van~Steenkiste, and J{\"u}rgen Schmidhuber.
\newblock On the binding problem in artificial neural networks.
\newblock {\em arXiv preprint arXiv:2012.05208}, 2020.

\bibitem{ha2016hypernetworks}
David Ha, Andrew Dai, and Quoc~V Le.
\newblock Hypernetworks.
\newblock {\em ICLR}, 2017.

\bibitem{kaiser2017learning}
{\L}ukasz Kaiser, Ofir Nachum, Aurko Roy, and Samy Bengio.
\newblock Learning to remember rare events.
\newblock {\em ICLR}, 2017.

\bibitem{kriete2013indirection}
Trenton Kriete, David~C Noelle, Jonathan~D Cohen, and Randall~C O'Reilly.
\newblock Indirection and symbol-like processing in the prefrontal cortex and
  basal ganglia.
\newblock {\em Proceedings of the National Academy of Sciences},
  110(41):16390--16395, 2013.

\bibitem{Krizhevsky2009LearningML}
A.~Krizhevsky.
\newblock Learning multiple layers of features from tiny images.
\newblock 2009.

\bibitem{le2020neural}
Hung Le, Truyen Tran, and Svetha Venkatesh.
\newblock Neural stored-program memory.
\newblock In {\em ICLR 2020: Proceedings of the 8th International Conference on
  Learning Representations}, 2020.

\bibitem{malsburg1994correlation}
Christoph von~der Malsburg.
\newblock The correlation theory of brain function.
\newblock In {\em Models of neural networks}, pages 95--119. Springer, 1994.

\bibitem{marcus2001algebraic}
Gary Marcus.
\newblock The algebraic mind, 2001.

\bibitem{mcinnes2018umap}
Leland McInnes, John Healy, and James Melville.
\newblock Umap: Uniform manifold approximation and projection for dimension
  reduction.
\newblock {\em arXiv preprint arXiv:1802.03426}, 2018.

\bibitem{mitchell2021abstraction}
Melanie Mitchell.
\newblock Abstraction and analogy-making in artificial intelligence.
\newblock {\em Annals of the New York Academy of Sciences}, 1505(1):79--101,
  2021.

\bibitem{Munkhdalai2019MetalearnedNM}
Tsendsuren Munkhdalai, Alessandro Sordoni, Tong Wang, and Adam Trischler.
\newblock Metalearned neural memory.
\newblock In {\em NeurIPS}, 2019.

\bibitem{parascandolo2018learning}
Giambattista Parascandolo, Niki Kilbertus, Mateo Rojas-Carulla, and Bernhard
  Sch{\"o}lkopf.
\newblock Learning independent causal mechanisms.
\newblock In {\em International Conference on Machine Learning}, pages
  4036--4044. PMLR, 2018.

\bibitem{rahaman2021dynamic}
Nasim Rahaman, Muhammad~Waleed Gondal, Shruti Joshi, Peter Gehler, Yoshua
  Bengio, Francesco Locatello, and Bernhard Sch{\"o}lkopf.
\newblock Dynamic inference with neural interpreters.
\newblock {\em Advances in Neural Information Processing Systems},
  34:10985--10998, 2021.

\bibitem{rowe2012cognitive}
Ellen~W Rowe, Cristin Miller, Lauren~A Ebenstein, and Dawna~F Thompson.
\newblock Cognitive predictors of reading and math achievement among gifted
  referrals.
\newblock {\em School Psychology Quarterly}, 27(3):144, 2012.

\bibitem{santoro2016meta}
Adam Santoro, Sergey Bartunov, Matthew Botvinick, Daan Wierstra, and Timothy
  Lillicrap.
\newblock Meta-learning with memory-augmented neural networks.
\newblock In {\em International conference on machine learning}, pages
  1842--1850. PMLR, 2016.

\bibitem{santoro2017simple}
Adam Santoro, David Raposo, David~G Barrett, Mateusz Malinowski, Razvan
  Pascanu, Peter Battaglia, and Timothy Lillicrap.
\newblock A simple neural network module for relational reasoning.
\newblock {\em Advances in neural information processing systems}, 30, 2017.

\bibitem{shanahan2020explicitly}
Murray Shanahan, Kyriacos Nikiforou, Antonia Creswell, Christos Kaplanis, David
  Barrett, and Marta Garnelo.
\newblock An explicitly relational neural network architecture.
\newblock In {\em International Conference on Machine Learning}, pages
  8593--8603. PMLR, 2020.

\bibitem{Vaswani2017AttentionIA}
Ashish Vaswani, Noam~M. Shazeer, Niki Parmar, Jakob Uszkoreit, Llion Jones,
  Aidan~N. Gomez, Lukasz Kaiser, and Illia Polosukhin.
\newblock Attention is all you need.
\newblock {\em NeurIPS}, abs/1706.03762, 2017.

\bibitem{vinyals2016matching}
Oriol Vinyals, Charles Blundell, Timothy Lillicrap, Daan Wierstra, et~al.
\newblock Matching networks for one shot learning.
\newblock {\em Advances in neural information processing systems}, 29, 2016.

\bibitem{webb2020learning}
Taylor Webb, Zachary Dulberg, Steven Frankland, Alexander Petrov, Randall
  O'Reilly, and Jonathan Cohen.
\newblock Learning representations that support extrapolation.
\newblock In {\em International conference on machine learning}, pages
  10136--10146. PMLR, 2020.

\bibitem{Webb2021EmergentST}
Taylor~Whittington Webb, Ishan Sinha, and Jonathan Cohen.
\newblock Emergent symbols through binding in external memory.
\newblock In {\em International Conference on Learning Representations}, 2020.

\bibitem{zhang2019raven}
Chi Zhang, Feng Gao, Baoxiong Jia, Yixin Zhu, and Song-Chun Zhu.
\newblock Raven: A dataset for relational and analogical visual reasoning.
\newblock In {\em Proceedings of the IEEE Conference on Computer Vision and
  Pattern Recognition (CVPR)}, 2019.

\end{thebibliography}

	\newpage
	
	\appendix

\section*{APPENDIX}

\section{\label{app:Transformations-details} Transformations Details}

We build IQ tasks with geometric transformation of 3 categories: affine,
non-linear and syntactic transformations. 

\subsubsection*{Affine transformations}

We consider 5 types of affine transformations: translation, rotation,
reflection, shear, and scaling. 
\begin{itemize}
\item \emph{Translation}: with translation vector $(i,j)$, where $i,j\in\{-9,-6,-3,0,3,6,9\}$.
\item \emph{Rotation}: with rotation angle $\alpha\in\{k\cdot15^{\circ}:k\in\{0,1,\ldots,23\}\}$.
\item \emph{Reflection}: horizontal or vertical reflection.
\item \emph{Shear}: with shear angles $(\alpha,\beta)$ where $\alpha,\beta\in\{-60^{\circ},-45^{\circ},-30^{\circ},-15^{\circ},0^{\circ},15^{\circ},30^{\circ},45^{\circ},60^{\circ}\}$.
\item \emph{Scaling}: with scale coefficient $s\in\{0.5,0.75,1,1.25\}$.
\end{itemize}

\subsubsection*{Non-linear transformations}

We consider 2 types of non-linear transformations: fisheye and horizontal
wave. 
\begin{itemize}
\item \emph{Fisheye}: given pixel $(x,y)$, the transformed pixel $(T(x),T(y))$
is given by $T(x)=x+(x-c_{x})\cdot d\cdot\sqrt{(x-c_{x})^{2}+(y-c_{y})^{2}}$
and $T(y)=y+(y-c_{y})\cdot d\cdot\sqrt{(x-c_{x})^{2}+(y-c_{y})^{2}}$,
where $(c_{x},c_{y})$ is the transformation center and $d$ is the
distortion factor. 
\item \emph{Horizontal wave}: given pixel $(x,y)$, the transformed pixel
$(T(x),T(y))$ is given by $T(x)=x$ and $T(y)=y+a\cos(fy)$, where
$a$ is the amplitude of cosine wave and $f$ is the frequency. 
\end{itemize}

\subsubsection*{Syntactic transformations}

We consider 2 types of syntactic transformations: black-white and
swap.
\begin{itemize}
\item \emph{Black-white}: the image is horizontally or vertically splitted
into 2 subimages (not necessarily of equal size). One subimage is
kept fixed, while the other one will be transformed $x\mapsto1-x$,
where $x$ is the pixel value. 
\item \emph{Swap}: the image is splitted into 4 equal subimages, which are
then permuted to achive the transformed image. 
\end{itemize}

\section{\label{app:theoretical-analysis} Principles for Designing the Hypothesis
Space $\mathcal{F}$ and the Function Composer $\phi$}

We aim to determine general principles for designing $\mathcal{F}$
and $\phi$. Suppose $\mu\colon\mathcal{X}\times\mathcal{Y}\to\mathcal{C}^{0}(\mathcal{X},\mathcal{Y})$,
where $\mathcal{C}^{0}(\mathcal{X},\mathcal{Y})$ is the space of
all continuous functions from $\mathcal{X}$ to $\mathcal{Y}$, be
the mapping that maps each input-output pair $(x,y)$ to the correct
function transforming $x$ to $y$. We further define a norm $\|.\|_{\mathcal{C}^{0}}$
on $\mathcal{C}^{0}(\mathcal{X},\mathcal{Y})$ determined by $\|f\|_{\mathcal{C}^{0}}={\displaystyle \sup_{x\in\mathcal{X}}\|f(x)\|_{\mathcal{Y}}}$,
where $\|.\|_{\mathcal{Y}}$ is an arbitrary norm on $\mathcal{Y}$.
Our goal is to find $\phi$ as the solution of the optimization problem: 

\vspace{-2mm}
\begin{equation}
{\displaystyle \text{Minimize}\quad\sum_{(x,y)}\|\mu_{x,y}-\phi_{x,y}\|_{\mathcal{C}^{0}}.}\label{eq:original_optimization_problem}
\end{equation}

\vspace{-2mm}

We hypothesize that the cardinality of the range $\mathcal{R}(\mu)$
of $\mu$ is much less than the number of data points (i.e. the number
of relations within the dataset is limited), and further supppose
$\mathcal{R}(\mu)=\{\mu_{1},\mu_{2},\ldots,\mu_{k}\}$, where $\mu_{i}$'s
are functions in $\mathcal{C}^{0}(\mathcal{X},\mathcal{Y})$. The
optimization problem in Eq.~((\ref{eq:original_optimization_problem}))
can be rewritten as: 

\vspace{-2mm}
\begin{equation}
\text{Minimize}\quad\sum_{i=1}^{k}\sum_{(x,y):\mu_{i}(x)=y}\|\mu_{i}-\phi_{x,y}\|_{\mathcal{C}^{0}}.\label{eq:sum_optimization_problem}
\end{equation}

\vspace{-2mm}

The optimization problem in Eq.~((\ref{eq:sum_optimization_problem}))
can be deduced to multiple optimization subproblems: 

\vspace{-2mm}

\begin{equation}
\text{Minimize}\quad\sum_{(x,y):\mu_{i}(x)=y}\|\mu_{i}-\phi_{x,y}\|_{\mathcal{C}^{0}},\qquad\forall i=1,2,\ldots,k.\label{eq:optimization_subproblems}
\end{equation}

\vspace{-2mm}

For each $i=1,2,\ldots,k$, let $\phi_{i}^{*}\in\mathcal{F}$ be the
function that best approximates $\mu_{i}$. By the triangle inequality,
we obtain 
\[
\|\mu_{i}-\phi_{x,y}\|_{\mathcal{C}^{0}}\le\|\mu_{i}-\phi_{i}^{*}\|_{\mathcal{C}^{0}}+\|\phi_{i}^{*}-\phi_{x,y}\|_{\mathcal{C}^{0}},\qquad\forall i=1,2,\ldots,k.
\]

Solving optimization problem Eq.~((\ref{eq:optimization_subproblems}))
might be difficult, so we instead consider an alternative optimization
problem 

\vspace{-2mm}
\begin{equation}
\text{Minimize}\quad\sum_{(x,y):\mu_{i}(x)=y}(\|\mu_{i}-\phi_{i}^{*}\|_{\mathcal{C}^{0}}+\|\phi_{i}^{*}-\phi_{x,y}\|_{\mathcal{C}^{0}}),\qquad\forall i=1,2,\ldots,k.\label{eq:triangle_optimization_subproblems}
\end{equation}

\vspace{-2mm}

We deduce following analysis after observing Eq.~((\ref{eq:triangle_optimization_subproblems})):
\begin{itemize}
\item The term $\|\phi_{i}^{*}-\phi_{x,y}\|_{\mathcal{C}^{0}}$ suggests
$\mathcal{R}(\phi)$ (the range of $\phi$) should not be too small
or too large, otherwise there may exist some $(x,y)$ such that $\phi_{x,y}$
is far away from $\phi_{i}^{*}$.
\item Since $\mathcal{R}(\phi)$ is contrained, so should be $\mathcal{F}$.
If $\mathcal{F}$ is too small, ${\displaystyle \|\mu_{i}-\phi_{i}^{*}\|_{\mathcal{C}^{0}}}$
may be large for some $i$; if $\mathcal{F}$ is too large, $\phi_{i}^{*}$
may be far away from $\mathcal{R}(\phi$), which leads to large $\|\phi_{i}^{*}-\phi_{x,y}\|_{\mathcal{C}^{0}}$.
\end{itemize}
With the above arguments, we suggest the following principles for
designing $\mathcal{F}$ and $\phi$:
\begin{enumerate}
\item $\mathcal{F}$ should be constrained by some prior knowledge of $\mu$.
For example, if we know $\mu$ is invertible, then $\mathcal{F}$
should also contain invertible functions only. 
\item $\phi$ should be determined on the fly in a meta-learning fashion
(associated with each input-output pair $(x,y)$) so that we can control
its complexity. 
\end{enumerate}

\section{\label{app:Training-setup} Training setup}

For ESBN, Transformer, RelationNet and PrediNet, we follow the same
settings as \cite{Webb2021EmergentST}, where all given images (including
examples and answer candidates) are treated as a sequence and passed
through a context normalization layer before being fed to the model.
For HyperNetwork, we also use the NICE backbone for fair comparisons
and maintain the key memories (but not the value memories) to compute
the weights; at each layer of the backbone, the attention weights
are computed as the output of an LSTM cell, where the input for LSTM
is the concatenation of the input and (pseudo-)output of current layer,
and the hidden states are taken from the LSTM cell of previous layer.
For FINE with NICE backbone, we use 4 NICE layers while using 2-layer
MLP for FINE with MLP backbone. We use 8-32 basis weight matrices
in the experiments. 

We use the Adam optimizer with no weight decay along with gradient
clipping with threshold 10 in all experiments. All tasks are trained
with 200-300 epochs. The training and testing batch sizes are 32 and
100, respectively. Feature vectors of images are of size 128.

\section{\label{app:More-results} More Results}

Table~\ref{tab:Omniglot-task-std} reports the full result table
with mean \& std on Omniglot dataset of single-transformation and
multi-affine-transformation task.

\begin{table}[th]
\begin{centering}
{\scriptsize{}}%
\begin{tabular}{l|>{\centering}p{0.7cm}|>{\centering}p{0.7cm}|>{\centering}p{0.7cm}|>{\centering}p{0.7cm}|>{\centering}p{0.7cm}|>{\centering}p{0.8cm}|>{\centering}p{0.8cm}|>{\centering}p{0.8cm}|>{\centering}p{0.8cm}|>{\centering}p{0.7cm}}
\cline{2-11} \cline{3-11} \cline{4-11} \cline{5-11} \cline{6-11} \cline{7-11} \cline{8-11} \cline{9-11} \cline{10-11} \cline{11-11} 
 & \multicolumn{9}{c|}{{\tiny{}Single-transformation}} & \multirow{3}{0.7cm}{\centering{}{\tiny{}Multi affine}}\tabularnewline
\cline{2-10} \cline{3-10} \cline{4-10} \cline{5-10} \cline{6-10} \cline{7-10} \cline{8-10} \cline{9-10} \cline{10-10} 
 & \multicolumn{5}{c|}{{\tiny{}Affine}} & \multicolumn{2}{c|}{{\tiny{}Non-linear}} & \multicolumn{2}{c|}{{\tiny{}Syntactic}} & \tabularnewline
\cline{2-10} \cline{3-10} \cline{4-10} \cline{5-10} \cline{6-10} \cline{7-10} \cline{8-10} \cline{9-10} \cline{10-10} 
 & \centering{}{\tiny{}Trans.} & \centering{}{\tiny{}Rot.} & \centering{}{\tiny{}Refl.} & \centering{}{\tiny{}Shear} & \centering{}{\tiny{}Scale} & \centering{}{\tiny{}Fish.} & \centering{}{\tiny{}H.Wave} & \centering{}{\tiny{}B\&W} & \centering{}{\tiny{}Swap} & \tabularnewline
\hline 
{\tiny{}RelationNet} & {\tiny{}27.1$\pm$0.4} & {\tiny{}26.2$\pm$0.3} & {\tiny{}25.5$\pm$0.4} & {\tiny{}27$\pm$0.5} & {\tiny{}27.5$\pm$0.4} & {\tiny{}26.1$\pm$0.7} & {\tiny{}30.2$\pm$6.9} & {\tiny{}49.7$\pm$29.1} & {\tiny{}26.0$\pm$2.2} & {\tiny{}25.3$\pm$0.2}\tabularnewline
{\tiny{}PrediNet} & {\tiny{}68.5$\pm$4.0} & {\tiny{}43.9$\pm$6.8} & {\tiny{}32.9$\pm$1.9} & {\tiny{}62.4$\pm$3.7} & {\tiny{}65.7$\pm$2.8} & {\tiny{}36.2$\pm$2.4} & {\tiny{}46.1$\pm$7.9} & {\tiny{}60.5$\pm$8.0} & {\tiny{}57.5$\pm$3.6} & {\tiny{}34.9$\pm$1.1}\tabularnewline
{\tiny{}HyperNet} & {\tiny{}88.9$\pm$1.0} & {\tiny{}62.0$\pm$2.9} & {\tiny{}94.0$\pm$1.9} & {\tiny{}74.5$\pm$1.4} & {\tiny{}81.8$\pm$1.1} & {\tiny{}63.2$\pm$2.0} & {\tiny{}80.4$\pm$1.0} & {\tiny{}88.6$\pm$1.4} & {\tiny{}90.1$\pm$1.0} & {\tiny{}54.0$\pm$4.1}\tabularnewline
{\tiny{}Transformer} & {\tiny{}89.5$\pm$1.0} & {\tiny{}64.8$\pm$1.5} & {\tiny{}44.3$\pm$0.9} & {\tiny{}86.3$\pm$4.1} & {\tiny{}84$\pm$0.9} & {\tiny{}41.4$\pm$11.6} & {\tiny{}91.0$\pm$11.8} & {\tiny{}97.6$\pm$0.4} & {\tiny{}49.9$\pm$18.5} & {\tiny{}59.4$\pm$6.0}\tabularnewline
{\tiny{}ESBN} & {\tiny{}79.8$\pm$0.6} & {\tiny{}58.6$\pm$1.0} & {\tiny{}50.1$\pm$0.3} & {\tiny{}83.4$\pm$1.6} & {\tiny{}84.5$\pm$1.2} & {\tiny{}67.1$\pm$0.8} & {\tiny{}86.4$\pm$1.0} & {\tiny{}90.5$\pm$4.1} & {\tiny{}71.6$\pm$2.7} & {\tiny{}63.1$\pm$0.5}\tabularnewline
\hline 
{\tiny{}$\ModelName$} & \textbf{\tiny{}94.3$\pm$0.4} & \textbf{\tiny{}77.6$\pm$0.7} & \textbf{\tiny{}95.1$\pm$1.0} & \textbf{\tiny{}87.2$\pm$0.3} & \textbf{\tiny{}86.6$\pm$0.4} & \textbf{\tiny{}78.5$\pm$0.7} & \textbf{\tiny{}95.9$\pm$0.4} & \textbf{\tiny{}98.4$\pm$0.3} & \textbf{\tiny{}96.2$\pm$0.2} & \textbf{\tiny{}69.1$\pm$0.6}\tabularnewline
\hline 
\end{tabular}{\scriptsize\par}
\par\end{centering}
\caption{\label{tab:Omniglot-task-std} Test accuracy (mean \& std) (\%) on
Omniglot dataset. }

\vspace{-3mm}

\end{table}

Table~\ref{tab:CIFAR100-single-transformation-task-std} reports
the full result table with mean \& std on CIFAR100 dataset of single-transformation
tasks.

\begin{table}[th]
\begin{centering}
\begin{tabular}{c|>{\centering}m{0.9cm}|>{\centering}m{0.9cm}|>{\centering}m{0.9cm}|>{\centering}m{0.9cm}|>{\centering}m{0.9cm}|>{\centering}m{0.9cm}|>{\centering}m{0.9cm}|>{\centering}m{0.9cm}|>{\centering}m{0.9cm}}
\cline{2-10} \cline{3-10} \cline{4-10} \cline{5-10} \cline{6-10} \cline{7-10} \cline{8-10} \cline{9-10} \cline{10-10} 
 & \multicolumn{5}{c|}{{\scriptsize{}Affine}} & \multicolumn{2}{c|}{{\scriptsize{}Non-linear}} & \multicolumn{2}{c}{{\scriptsize{}Syntactic}}\tabularnewline
\cline{2-10} \cline{3-10} \cline{4-10} \cline{5-10} \cline{6-10} \cline{7-10} \cline{8-10} \cline{9-10} \cline{10-10} 
 & {\scriptsize{}Trans.} & {\scriptsize{}Rot.} & {\scriptsize{}Refl.} & {\scriptsize{}Shear} & {\scriptsize{}Scale} & {\scriptsize{}Fish.} & {\scriptsize{}H.Wave } & {\scriptsize{}B\&W} & {\scriptsize{}Swap}\tabularnewline
\hline 
{\scriptsize{}RelationNet} & {\scriptsize{}59.9$\pm$11.2} & {\scriptsize{}49.6$\pm$5.8} & {\scriptsize{}29.9$\pm$1.0} & {\scriptsize{}45.3$\pm$5.1} & {\scriptsize{}66.2$\pm$1.3} & {\scriptsize{}28.7$\pm$1.0} & {\scriptsize{}39.5$\pm$6.9} & {\scriptsize{}26.2$\pm$1.4} & {\scriptsize{}29.7$\pm$0.9}\tabularnewline
{\scriptsize{}PrediNet} & {\scriptsize{}72.4$\pm$4.6} & {\scriptsize{}65.6$\pm$5.0} & {\scriptsize{}40.6$\pm$2.0} & {\scriptsize{}74.3$\pm$4.8} & {\scriptsize{}76.1$\pm$3.6} & {\scriptsize{}37.1$\pm$1.2} & {\scriptsize{}53.9$\pm$8.1} & {\scriptsize{}32.7$\pm$1.8} & {\scriptsize{}39.6$\pm$1.3}\tabularnewline
{\scriptsize{}HyperNet} & {\scriptsize{}94.8$\pm$1.1} & {\scriptsize{}86.8$\pm$1.3} & {\scriptsize{}46.6$\pm$0.5} & {\scriptsize{}91.3$\pm$0.9} & {\scriptsize{}85.2$\pm$1.2} & {\scriptsize{}46.8$\pm$0.7} & {\scriptsize{}80.5$\pm$4.6} & {\scriptsize{}47.8$\pm$0.9} & {\scriptsize{}46$\pm$0.8}\tabularnewline
{\scriptsize{}Transformer} & {\scriptsize{}98.4$\pm$1.1} & {\scriptsize{}86.3$\pm$3.8} & {\scriptsize{}47.5$\pm$1.1} & {\scriptsize{}95.4$\pm$1.4} & {\scriptsize{}84.9$\pm$1.2} & {\scriptsize{}47.2$\pm$1.0} & {\scriptsize{}95.1$\pm$1.8} & {\scriptsize{}51.6$\pm$14.3} & {\scriptsize{}47.6$\pm$0.8}\tabularnewline
{\scriptsize{}ESBN} & {\scriptsize{}96.6$\pm$0.7} & {\scriptsize{}81.9$\pm$1.1} & {\scriptsize{}50.6$\pm$0.4} & {\scriptsize{}90.1$\pm$0.7} & {\scriptsize{}81.5$\pm$0.9} & {\scriptsize{}57.7$\pm$1.3} & {\scriptsize{}95.7$\pm$0.7} & {\scriptsize{}68.8$\pm$6.0} & {\scriptsize{}50.5$\pm$0.5}\tabularnewline
\hline 
{\scriptsize{}$\ModelName$} & \textbf{\scriptsize{}99.2$\pm$0.1} & \textbf{\scriptsize{}91.3$\pm$0.2} & \textbf{\scriptsize{}80.6$\pm$14.5} & \textbf{\scriptsize{}95.6$\pm$0.5} & \textbf{\scriptsize{}87$\pm$0.2} & \textbf{\scriptsize{}76.8$\pm$1.3} & \textbf{\scriptsize{}98.3$\pm$0.4} & \textbf{\scriptsize{}89.1$\pm$0.7} & \textbf{\scriptsize{}51.6$\pm$1.7}\tabularnewline
\hline 
\end{tabular}
\par\end{centering}
\caption{\label{tab:CIFAR100-single-transformation-task-std} Test accuracy
(\%) on CIFAR100 dataset of single-transformation tasks.}

\vspace{-3mm}
\end{table}

\end{document}